\def\year{2021}\relax
\documentclass[letterpaper]{article} 
\usepackage{aaai21}  
\usepackage{times}  
\usepackage{helvet} 
\usepackage{courier}  
\usepackage[hyphens]{url}  
\usepackage{graphicx} 
\usepackage{makecell}
\usepackage{natbib}  
\usepackage{caption} 
\title{Fashion Focus: Multi-modal Retrieval System for Video Commodity Localization in E-commerce}
\author{
Yanhao Zhang, Qiang Wang, Pan Pan, Yun Zheng, Cheng Da, Siyang Sun and Yinghui Xu
}
\affiliations{
    Machine Intelligence Technology Lab, Alibaba Group \\
\{yanhao.zyh, qishi.wq, panpan.pp, zhengyun.zy, dacheng.dc, siyang.ssy, renji.xyh\}@alibaba-inc.com

}

\begin{document}
\maketitle

\begin{abstract}
Nowadays, live-stream and short video shopping in E-commerce have grown exponentially. However, the sellers are required to manually match images of the selling products to the timestamp of exhibition in the untrimmed video, resulting in a complicated process. To solve the problem, we present an innovative demonstration of multi-modal retrieval system called ``Fashion Focus'', which enables to exactly localize the product images in the online video as the focuses. Different modality contributes to the community localization, including visual content, linguistic features and interaction context are jointly investigated via presented multi-modal learning. Our system employs two procedures for analysis, including video content structuring and multi-modal retrieval, to automatically achieve accurate video-to-shop matching. Fashion Focus presents a unified framework that can orientate the consumers towards relevant product exhibitions during watching videos and help the sellers to effectively deliver the products over search and recommendation.
\end{abstract}

\section{Introduction}
Commodity delivery on live-stream and short videos becomes increasingly popular on E-commerce platform. Consumers purchase their favorite products during the process of watching videos. During the live-stream, the sellers often display, exhibit and introduce hundreds of products. If the customers intend to buy the items that the seller is explaining, they have to select them manually from the list associated with the live-stream, which greatly affects the purchasing efficiency and consuming experience. Thus, it would be more valuable to automatically identify the currently explained items according to the visual and explaining content, and recommend the corresponding purchase link of the products to the consumers. The shopping experience of users will be greatly improved during the video watching.
\begin{figure}[t]
\begin{center}
\includegraphics[width=1.0\linewidth]{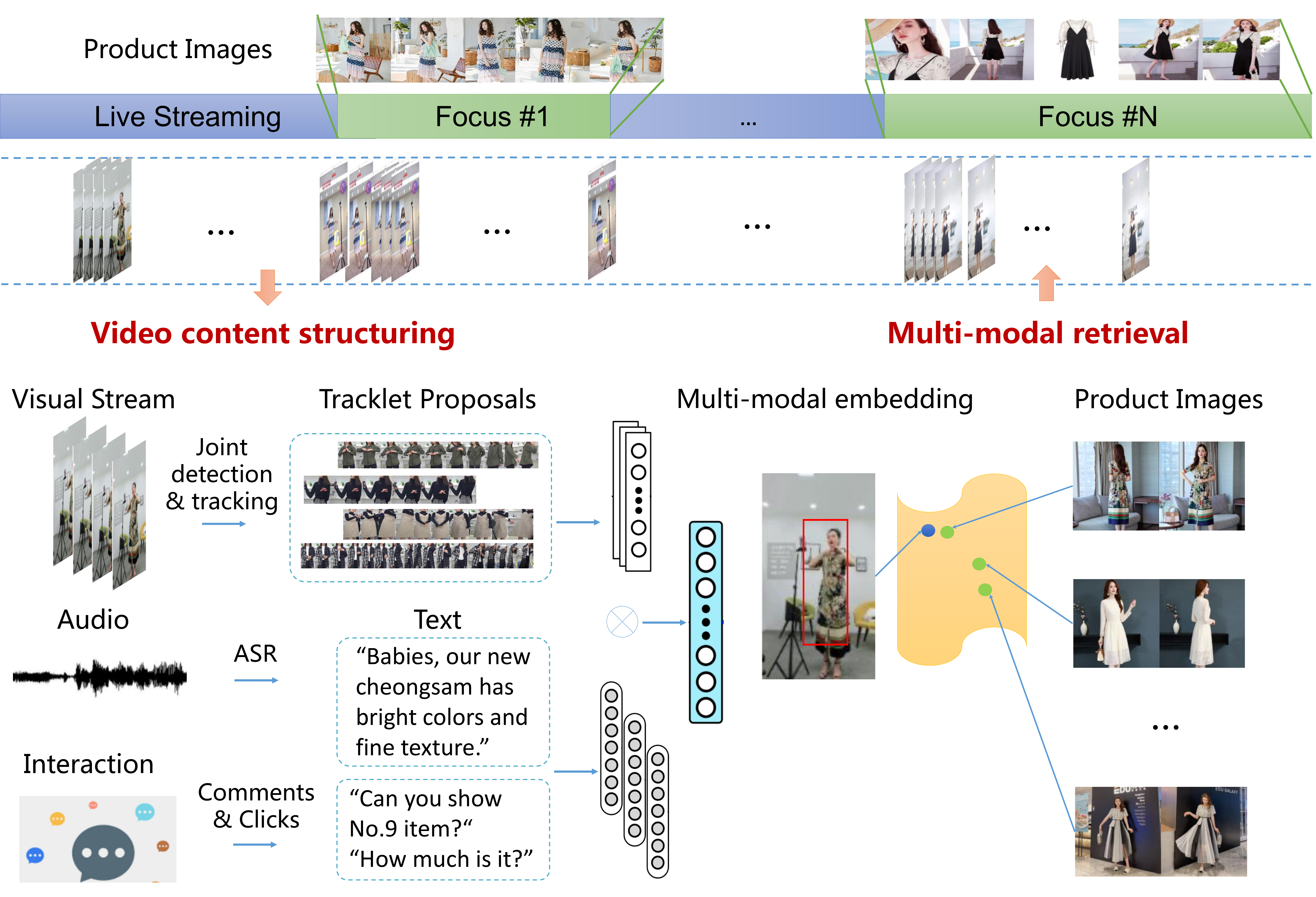}
\caption{Overview of our system. Given untrimmed video, our system can simultaneously generate tracklet proposals and combine different modalities (visual and linguistic data) to learn multi-modal embedding for retrieval. We associate start and end time of the exhibition during the video with the product images as the focus to deliver.}
\label{fig:framework}
\end{center}
\end{figure}

Unlike traditional video-to-shop retrieval, commodity localization in live-stream is more challenging due to various viewpoints of exhibition and distraction of fore-background products. Therefore, previous visual search approaches~\cite{zhang2018visual} would not work well in such scenarios. Recently, many works are proposed to search identical fashion images from videos in cross-domain manner~\cite{ijcai_ZhaoYLWLYW20,cvpr_ChengWLH17}. However, accurately recognizing the fashion products in live-stream still faces the limitation of insufficient data. Fortunately, live-stream videos are usually accompanied with audio explanations and interaction comments, which complement rich contextual information with weak annotations. Moreover, automatically extracting features from multiple modalities by visual CNN models, Automatic Speech Recognition(ASR) and NLP tools can be useful for video-to-shop matching.

In this demo, we present a system called ``Fashion Focus'', which enables accurate localization of commodity in live-steam videos. Two essential tasks are solved, tracklet proposals generation and multi-modal embedding. In contrast to previous works, we consider the multi-modal learning paradigm, which combines visual and linguistic features for commodity understanding and representation learning.

\begin{figure}[t]
\begin{center}
\includegraphics[width=1.0\linewidth]{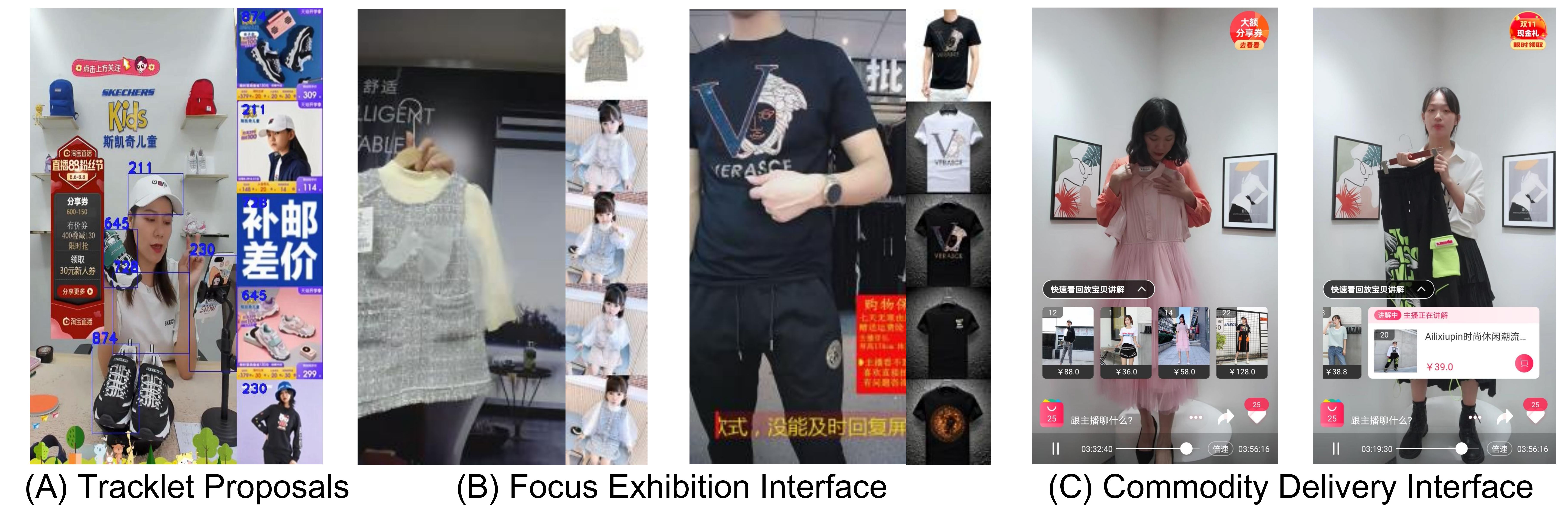}
\caption{Fashion Focus examples and system interface.}
\label{fig:system}
\end{center}
\end{figure}
\section{System Framework}
In proposed demo system, image list of products and live-stream video are given as input, the matched products with timestamp are localized as output. Figure~\ref{fig:framework} shows the work flow of our system based on Taobaolive platform\footnote{https://taobaolive.taobao.com/}. The system first extracts visual proposals, speech audio and linguistic data of surrounding comments, and then performs efficient retrieval through a unified embedded learning. Therefore, it mainly involves two procedures: video content structuring and multi-modal retrieval. A streaming engine is developed to provide real-time services for tremendous videos.

\subsection{Video Content Structuring}
Video content structuring aims to locate instances in untrimmed videos with both categories and consistent object tracks. The implementation is a composite task that requires the joint detection and tracking of objects in videos. We utilize an efficient one-stage video detector~\cite{fairmot} to generate tracklet proposals. We jointly learn a multi-task tracker with detection and appearance embedding branches, which allows to obtain high accuracy of detection and tracking with high speed. The detection model is extended with a tracking head and employs a DLA-34~\cite{dla} backbone. Although convincing detection accuracy can be achieved, recognition at scale presents serious long tail problem. Our module uses a hierarchical view of object classification that allows us to alleviate the vanishing-classification problem and reduce the category ambiguity.

\subsection{Efficient Multi-Modal Retrieval in Video}

%

Besides video, audio and captions are pervasive in search engines and social networks. To guarantee retrieval quality and computation efficiency, our framework presents a compact retrieval solution with multi-modal information using visual and language representation, which improves retrieval quality by tracklet-to-image metric learning with triplet constraint~\cite{cvpr_ChengWLH17}. In addition to tracklet feature, we convert audio into text by ASR and use LXMERT~\cite{LXMERT_TanB19} model for visual-audio embedding extraction, which has already shown the superior performance over NLP field. Further, we mine the comments and headlines to enhance the linguistic data. The above multi-modal information is fused and mapped to a compact embedding for retrieval with the commodity images. The main contribution lies in jointly learning visual embeddings and the superior linguistic features, which greatly increase the recall of products.

In addition, the whole system needs to provide real-time online services for tremendous videos. We develop a streaming engine of decoding, embedding and retrieval. Figure~\ref{fig:system} illustrates the visualization of (A) tracklet proposals, (B) focus exhibition interface for seller side and (C) commodity delivery interface for consumer side.
\section{Experiments}
We investigate the contributions of several related components in ``Fashion Focus'' and implement multiple architectures of related components. The experiments are conducted on the dataset containing 1K live-steam videos of six categories (clothe, bag, etc) with duration about 5-7 hours. The recall of identical matched products is adopted as evaluation metric. The contributed components in ``Fashion Focus'' is composed of visual modality with image CNN feature (Vis\_Img), visual modality with tracklet feature (Vis\_Trk), and multiple modalities with tracklet feature (Mlt\_Trk). We notice that Mlt\_Trk  with multi-modal embedding achieves the best result on all the categories in Table~\ref{pailitao_exper}, which verify the superior performance for video-to-shop matching.
\begin{table}[t]
\begin{center}
\LARGE
\resizebox{1.0\linewidth}{!}{%
\begin{tabular}{l|c|c|c|c|c|c|c}
\Xhline{1.0pt}
Method& clothing & shoe & bag & snack & bottle & beauty  & mean \\
\Xhline{1.0pt}
Vis\_Img & 16.4\% & 15.1\% & 13.7\% & 7.1\% & 10.4\% & 12.7\% & 12.5\% \\
\hline
Vis\_Trk & 19.3\% & 17.1\% & 16.4\% & 8.5\% & 12.4\% & 14.7\% & 14.7\% \\
\hline
\textbf{Mlt\_Trk} & 23.6\% & 19.2\% & 18.3\% & 10.4\% & 13.1\% & 16.1\% & \textbf{16.8\%}\\
\Xhline{1.0pt}
\end{tabular}}
\end{center}
\caption{The recall comparison of commodity localization between related components in Fashion Focus.}
\label{pailitao_exper}
\end{table}
\section{Conclusion}
In this demo, Fashion Focus provides exact commodity localization for live-stream video. The work included two aspects: 1) video content structuring based on joint detection and tracking, 2) learning multi-modal embedding for effective and efficient retrieval. Our system achieves favourable performance in connecting the video and product, which provides satisfying watch and buy experience.
\bibliography{focus}

\end{document}